\pdfoutput=1

\documentclass[11pt]{article}

\usepackage[final]{acl}

\usepackage{times}
\usepackage{latexsym}

\usepackage[T1]{fontenc}

\usepackage[utf8]{inputenc}

\usepackage{microtype}

\usepackage{inconsolata}

\usepackage{graphicx}
\usepackage{booktabs}
\usepackage{subcaption}
\usepackage{array}     
\usepackage{siunitx}  
\usepackage{amsmath}   
\usepackage{multirow}
\usepackage{tabularx}
\usepackage{adjustbox}
\usepackage{makecell}
\usepackage{url}

\title{Who is in the Spotlight: The Hidden Bias Undermining Multimodal Retrieval-Augmented Generation}

\author{
    Jiayu Yao$^{1,2}$\thanks{~~Work performed primarily as a visiting student at the Institute of Computing Technology, Chinese Academy of Sciences; currently a student at Beijing University of Posts and Telecommunications (BUPT).} 
    \thanks{~~Authors from affiliation $^{2}$ are also affiliated with: Key Laboratory of Network Data Science and Technology, ICT, CAS; State Key Laboratory of AI Safety; University of Chinese Academy of Sciences.} \quad
    Shenghua Liu$^{2}$\footnotemark[2] \thanks{~~Corresponding author.}\quad
    Yiwei Wang$^{3}$ \quad
    Lingrui Mei$^{2}$\footnotemark[2] \quad \\
    \textbf{Baolong Bi}$^{2}$\footnotemark[2] \quad
    \textbf{Yuyao Ge}$^{2}$\footnotemark[2] \quad
    \textbf{Zhecheng Li}$^{4}$ \quad
    \textbf{Xueqi Cheng}$^{2}$ \\
    $^1$Beijing University of Posts and Telecommunications \\
    $^2$Institute of Computing Technology, Chinese Academy of Sciences \\
    $^3$University of California, Merced \\
    $^4$University of California, San Diego \\
    {\small \texttt{yaojiayu@bupt.edu.cn}, \texttt{\{liushenghua, meilingrui22, geyuyao24z\}@ict.ac.cn}}
}

\makeatletter
\renewcommand\@makefntext[1]{%
  \noindent\makebox[-1pt][l]{\@makefnmark\hspace{0em}}#1}
\makeatother

\begin{document}
\maketitle
\begin{abstract}
Multimodal Retrieval-Augmented Generation (RAG) systems have become essential in knowledge-intensive and open-domain tasks. As retrieval complexity increases, ensuring the robustness of these systems is critical. However, current RAG models are highly sensitive to the order in which evidence is presented, often resulting in unstable performance and biased reasoning, particularly as the number of retrieved items or modality diversity grows. This raises a central question: \textit{How does the position of retrieved evidence affect multimodal RAG performance?} To answer this, we present the first comprehensive study of position bias in multimodal RAG systems. Through controlled experiments across text-only, image-only, and mixed-modality tasks, we observe a consistent U-shaped accuracy curve with respect to evidence position. To quantify this bias, we introduce the Position Sensitivity Index ($PSI_p$) and develop a visualization framework to trace attention allocation patterns across decoder layers. Our results reveal that multimodal interactions intensify position bias compared to unimodal settings, and that this bias increases logarithmically with retrieval range. These findings offer both theoretical and empirical foundations for position-aware analysis in RAG, highlighting the need for evidence reordering or debiasing strategies to build more reliable and equitable generation systems.
\end{abstract}

\section{Introduction}
The growing demand for multimodal interaction has driven the development of multimodal Retrieval-Augmented Generation (RAG) systems, which integrate heterogeneous data sources (text, images, audio) to achieve comprehensive information understanding \cite{abootorabi2025askmodalitycomprehensivesurvey}. This technological advancement has enabled breakthroughs across diverse domains: academic research leverages frameworks like Taichu-mRAG, OmniSearch, VARAG \cite{faysse2024colpaliefficientdocumentretrieval} and GraphRAG \cite{edge2025localglobalgraphrag} for knowledge discovery, while industrial applications employ DocPrompting \cite{zhou23docprompting}, UniFashion \cite{Zhao2024UniFashionAU}, RAG-Driver \cite{yuan2024rag} and Img2Loc \cite{zhou2024img2loc} for real-world problem solving. In professional fields, systems like RULE \cite{xia-etal-2024-rule} and MMed \cite{xia2024mmedrag} enhance medical diagnostics. In other scenarios, SoccerRAG\cite{strand2024soccerragmultimodalsoccerinformation} and MMRA\cite{wu2024mmra} research demonstrates the social computing and entertainment applications. These successes highlight multimodal RAG's potential in open-domain question answering and knowledge-intensive scenarios.
\begin{figure*}[t]
    \centering
    \includegraphics[width=1\linewidth]{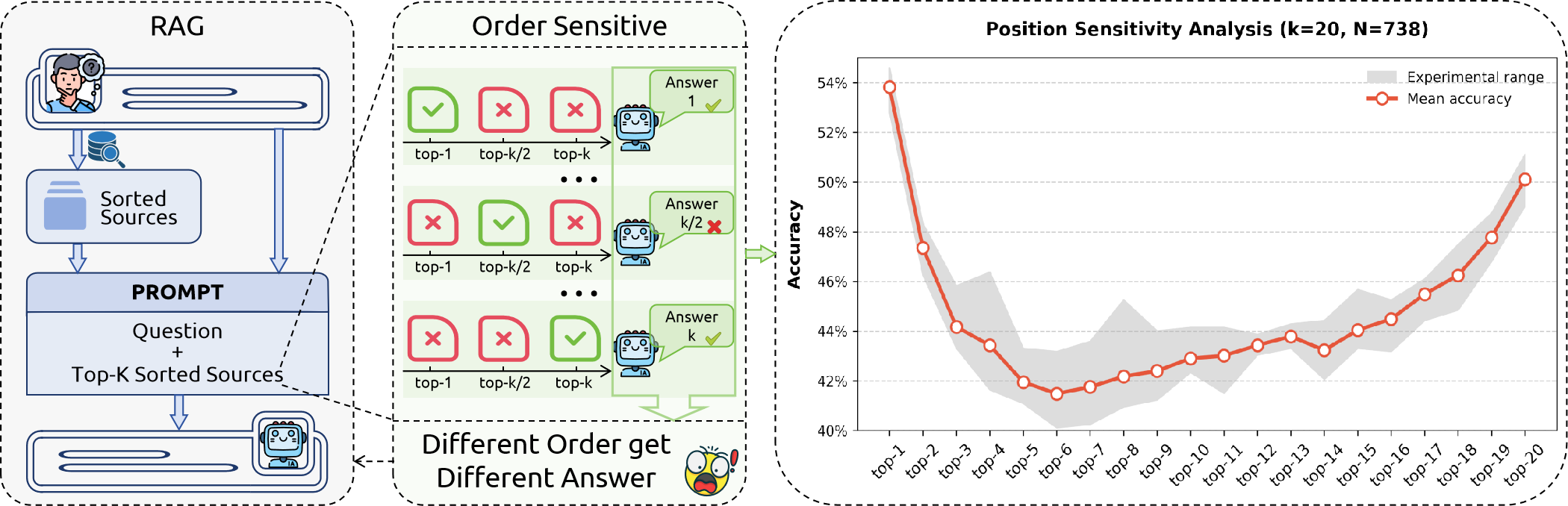} 
    \caption {{\bf Left }An example of RAG for open question answering, where the prompt contains the question and $top-k$ sorted documents. {\bf Mid }Different positions of the search results may lead to different answers. {\bf Right }Accuracy analysis when correct document ranks at position $i$ ($i\in k, k=20$) - gray blocks display 10 random experiments, red line indicates mean accuracy.}
    \label{fig:pipline} 
\end{figure*}

However, the reliability of multimodal RAG systems faces critical challenges as applications expand. Current systems exhibit vulnerability when faced with complex retrieval problems: excessive or insufficient retrieved content often induces hallucination. Ensuring that generated outputs faithfully adhere to the provided context is crucial for overall system robustness \cite{bi2024context}, yet even with optimal corpus selection, the positional arrangement of retrieved results significantly impacts answer reliability when fed to generation models. This instability aligns with emerging researches on systematic position bias in contemporary Large Language Models (LLMs) and Vision-Language Models (VLMs) \cite{tan2024ordermattersexploringorder, zhang-etal-2024-instruct}. These models disproportionately focus on the start and end positions of input sequences while neglecting middle content, a phenomenon we term "middle-loss". However, the existing research presents two critical gaps. First, current studies primarily address single-modal RAG systems (e.g. text-based), lacking systematic investigation of multimodal scenarios. Second, conventional evaluation metrics (e.g. NDCG, MRR) fail to quantify positional sensitivity, with no interpretable framework established for bias analysis in multimodal contexts.

To address these limitations, we obtained the following research results through controlled experiments and interpretability analysis:
\begin{itemize}
    \item We show that the multimodal RAG system has position bias (Figure \ref{fig:pipline}), and in text, image and mixed-modality scenarios, the generate model will give different degrees of attention to the evidence at different positions, forming a U-shaped accuracy curve.
    \item We propose a position sensitivity index $PSI_p$ to quantify the amplitude of position bias. Through benchmarking across modalities and retrieval scales ($k$), we demonstrate that cross-modal interactions and larger retrieval sizes may amplify bias severity. Furthermore, we conduct mathematical modeling between the retrieval scale and the bias amplitude, providing guidance for balancing the retrieval range and robustness in deployment.    
    \item Through systematic attention visualization analysis, we experimentally validate the above conclusions and uncover layer-specific sparsity patterns in cross-modal attention across decoder hierarchies.
\end{itemize}

\section{Position Bias}
\begin{figure*}[t]
\begin{center}
\includegraphics[width=1\linewidth]{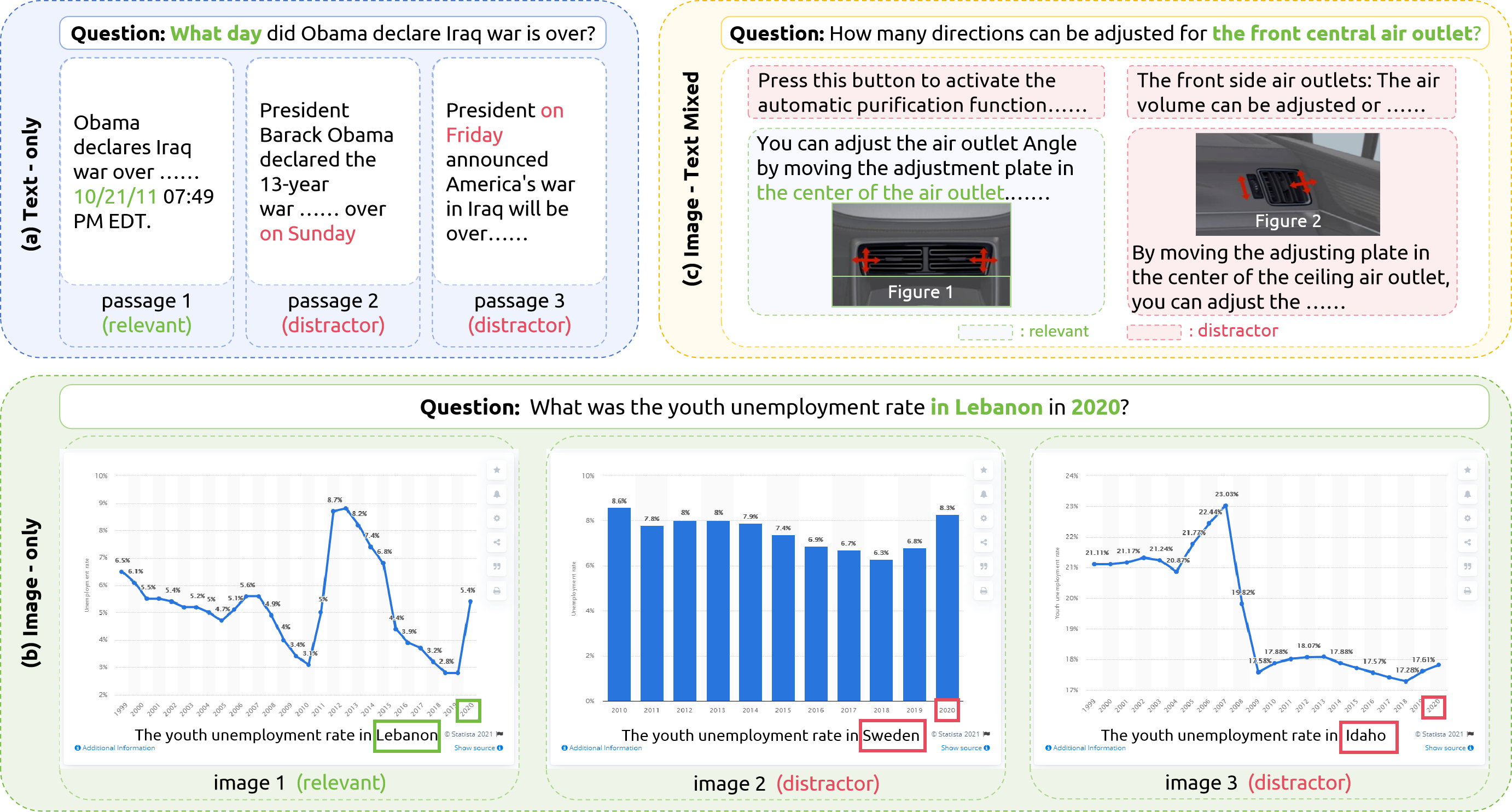} 
\end{center}
\caption{Order sensitivity evaluation tasks in three modality configurations: (a) Text-only, (b) Image-only, (c) Mixed-modality.}
\label{fig:dataset} 
\end{figure*}

To better understand how the position of retrieved evidences affect reasoning in multimodal RAG systems, we begin with a set of controlled experiments designed to probe position bias across diverse input modalities. This section presents both the experimental setup and key empirical findings, revealing a consistent pattern of position-induced performance fluctuation. By systematically perturbing the position of gold evidence while holding content constant, we are able to isolate and quantify the impact of sequence order on model behavior across text-only, image-only, and mixed-modality contexts.
To simulate real-world open-domain question answering and knowledge-intensive scenarios, we conducted controlled experiments across three modality configurations: text-only, image-only, and image-text mixed. As depicted in Figure \ref{fig:dataset}, our framework emulates practical RAG workflows where systems process multiple evidence documents (text passages, charts, or multimodal pairs) alongside user queries to generate answers. Through systematic dataset perturbation, we preserve semantic content while manipulating gold evidence positions relative to distractors, with 10 randomized experiments per configuration to ensure statistical robustness. Through this design, we isolate positional effects from content biases, enabling precise characterization of context ordering sensitivity.

\subsection{Benchmark}
\paragraph{Data}
For text-based QA simulation, we leverage the MS-MARCO passage ranking benchmark \cite{DBLP:journals/corr/NguyenRSGTMD16}, constructing triplets comprising a query, one gold passage containing the answer, and two topically relevant but non-answer distractor passages. Gold passage permutation across three positions (Top/Mid/End) mimics real-world multi-document retrieval scenarios. 

In VQA tasks, we use ChartQA \cite{masry-etal-2022-chartqa} to create chart triplets (one answer-containing gold chart and two distractors) with identical positional variations to evaluate pure image processing. For image-text mixed-modality evaluation, we use the VEGA dataset \cite{zhou2024vegalearninginterleavedimagetext}. Each sample in this dataset includes a question, several image-text pairs, and a golden answer based on a specific pair. In our experiment, for each query, we selected the correct image-text pairs and randomly chose two pairs of unrelated images and paragraphs as interference pairs. Then, we set the same position transformation to evaluate whether the multimodal RAG model shows sequence sensitivity when reasoning about interlaced visual and textual evidence.

\paragraph{Models}
Our evaluation protocol preserves validity through frozen model parameters and deactivated fine-tuning functions, ensuring comparisons reflect inherent architectural biases rather than training artifacts. For the retrieval stage, we implement cross-modal retrieval using the VisRAG-Ret \cite{yu2025visragvisionbasedretrievalaugmentedgeneration}. In the generation stage, we systematically evaluate three cutting-edge open-source instruction-tuned models (Qwen2-VL-7B-Instruct \cite{Qwen2VL}, Llama-3.2-11B-Vision-Instruct \cite{llama3modelcard}, MINICPM-v2.6 \cite{minicpmv26}) and the closed-source GPT-4o \cite{gpt4o}, which support multi-image input integration and cross-modal attention interactions.

\subsection{Empirical Patterns of Positional Bias}
Our experiments are conducted on a high-performance computing system with 8×NVIDIA A100 GPUs (80GB memory each), an Intel Xeon Platinum 8336C processor (128 threads @3.5GHz), and 2TB RAM. The software environment runs on Ubuntu 20.04 with CUDA 12.4 for GPU acceleration.
\begin{table*}[htbp]
\centering
\small
\renewcommand{\arraystretch}{1.25}
\adjustbox{max width=\linewidth}{ 
\begin{tabular}{l c >{\centering\arraybackslash}m{2.5cm} >{\centering\arraybackslash}m{2.5cm} >{\centering\arraybackslash}m{2.5cm} >{\centering\arraybackslash}m{2.5cm}}
\toprule
\midrule
\textbf{Modality} & \textbf{Location} & \textbf{MINICPM-v2.6} & \textbf{Qwen2-VL-7B} & \textbf{Llama-3.2-11B-Vision} & \textbf{GPT-4o} \\
\midrule
\multirow{3}{*}{\textbf{Text-only}} 
& Top & $0.5321{\scriptstyle ( \pm 0.007 )}$ & $0.4182{\scriptstyle ( \pm 0.009 )}$ & $0.3199{\scriptstyle ( \pm 0.012 )}$ & $0.5083{\scriptstyle ( \pm 0.010 )}$ \\
& Mid & $0.4963{\scriptstyle ( \pm 0.011 )}$ & $0.3810{\scriptstyle ( \pm 0.010 )}$ & $0.2992{\scriptstyle ( \pm 0.009 )}$ & $0.4513{\scriptstyle ( \pm 0.008 )}$ \\
& End & $0.5075{\scriptstyle ( \pm 0.010 )}$ & $0.3882{\scriptstyle ( \pm 0.008 )}$ & $0.3095{\scriptstyle ( \pm 0.011 )}$ & $0.4679{\scriptstyle ( \pm 0.014 )}$ \\
\midrule
\multirow{3}{*}{\textbf{Image-only}} 
& Top & $0.5911{\scriptstyle ( \pm 0.010 )}$ & $0.4859{\scriptstyle ( \pm 0.011 )}$ & $0.2256{\scriptstyle ( \pm 0.009 )}$ & $0.7333{\scriptstyle ( \pm 0.007 )}$ \\
& Mid & $0.5506{\scriptstyle ( \pm 0.009 )}$ & $0.4790{\scriptstyle ( \pm 0.008 )}$ & $0.2103{\scriptstyle ( \pm 0.013 )}$ & $0.7059{\scriptstyle ( \pm 0.012 )}$ \\
& End & $0.5593{\scriptstyle ( \pm 0.007 )}$ & $0.5114{\scriptstyle ( \pm 0.010 )}$ & $0.2019{\scriptstyle ( \pm 0.010 )}$ & $0.8125{\scriptstyle ( \pm 0.008 )}$ \\
\midrule
\multirow{3}{*}{\makecell{\textbf{Image-Text Mixed}s}} 
& Top & $0.5057{\scriptstyle ( \pm 0.009 )}$ & $0.3891{\scriptstyle ( \pm 0.009 )}$ & $0.2496{\scriptstyle ( \pm 0.011 )}$ & $0.7557{\scriptstyle ( \pm 0.010 )}$ \\
& Mid & $0.4599{\scriptstyle ( \pm 0.010 )}$ & $0.3544{\scriptstyle ( \pm 0.008 )}$ & $0.2272{\scriptstyle ( \pm 0.010 )}$ & $0.7021{\scriptstyle ( \pm 0.008 )}$ \\
& End & $0.4671{\scriptstyle ( \pm 0.013 )}$ & $0.3923{\scriptstyle ( \pm 0.013 )}$ & $0.2202{\scriptstyle ( \pm 0.009 )}$ & $0.7462{\scriptstyle ( \pm 0.009 )}$ \\
\bottomrule
\end{tabular}
}
\caption{Overall generation performance in accuracy (\%) across different input modalities and evidence positions, with standard deviations in parentheses.}
\label{tab:modalitiesExp}
\end{table*}

Our experiment reveal the systematically existing position bias patterns in multimodal scenarios through controlled simulation of RAG scenarios. As shown in the Table \ref{tab:modalitiesExp}, the average results based on ten randomized experiments indicate: When the key evidence documents are located in the middle of the input sequence, the generative modelability generally decreases, while the top and end positions maintain a relatively high accuracy rate, forming a typical U-shaped performance curve. This phenomenon stably exists in text (MS-MARCO), image (ChartQA), and mixed-modality (VEGA) scenarios. Among them, GPT-4o shows a significant advantage in the recovery ability at the tail position (the accuracy at the tail of the image task is 11\% higher than that in the middle). It can be seen from the results that the performance of MINICPM-v2.6 and Qwen2-VL-7B-Instruct is relatively weak in multiple tasks, which is mainly attributed to their deficiencies in following instructions. 

The high consistency of the position bias pattern in cross-modal scenarios confirms the order sensitivity feature that is prevalent in the multimodal RAG architecture, and this feature is independent of the specific evidence mode. Subsequently, based on this phenomenon, we will further quantify the amplitude of position bias and analyze its mechanism.

\section{Evaluation and Analysis}
\subsection{Quantification of Bias Amplitude}
To quantify the intensity of the position bias problem of large models under three modalities, we define the position sensitivity as:
\begin{equation}
\begin{aligned}
PSI_p \; = \; \frac{1}{p}\sum_{i\in \mathcal{T}_p} A_i \; - \; \frac{1}{p}\sum_{j\in \mathcal{B}_p} A_j.
\end{aligned}
\end{equation}

Among them, $A_i$represents the accuracy rate when the evidence is placed at the $i$th position; $\mathcal{T}_p$ and $\mathcal{B}_p$ respectively represent the set of $p$ position indexes with the highest and lowest accuracy rates. When p=1, p=1 and the highest/lowest positions are respectively at the beginning, the end and the middle, this indicator is denoted as $PSI$.

Based on the accuracy data in the above Table \ref{tab:modalitiesExp}, we calculated the $PSI$ values for MS-MARCO (text), ChartQA (image), and VEGA (image-text mixed) scenarios, respectively. The results are summarized in Table \ref{tab:modalitiesPSI}.

\begin{table}[htbp]
\centering
\renewcommand{\arraystretch}{1.15}
\adjustbox{max width=\columnwidth}{
\begin{tabular}{l c >{\centering\arraybackslash}m{1.7cm} >{\centering\arraybackslash}m{1.7cm} >{\centering\arraybackslash}m{1.7cm} >{\centering\arraybackslash}m{1.7cm}}
\toprule
\midrule
\textbf{Modality / Model} & \textbf{MiniCPM} & \textbf{Qwen2-VL} & \textbf{Llama-3.2} & \textbf{GPT-4o} \\
\midrule
\textbf{Text-only} 
& 0.0358 & 0.0372 & 0.0207 & 0.0570 \\
\midrule
\textbf{Image-only} 
& 0.0405 & 0.0324 & 0.0237 & 0.1066 \\
\midrule
\textbf{Image-Text Mixed} 
& 0.0458 & 0.0379 & 0.0294 & 0.0536 \\
\bottomrule
\end{tabular}
}
\caption{Comparison of Position sensitivity $PSI$ of each model in the three modalities.}
\label{tab:modalitiesPSI}
\vspace{-1em} 
\end{table}

It can be seen from the Table \ref{tab:modalitiesPSI} that there are universal differences in the position sensitivity of the model under different modal scenarios. In the text-only scenario, the $PSI$values of all models are between 0.020 and 0.050, indicating that the influence of the positions of text evidence on the generation performance is relatively balanced and moderate. Secondly, the image-only task amplified the sensitivity of most models. The $PSI$ of GPT-4o soared to 0.1066, which was 86.9\% higher than the text-only baseline. Paradoxically, Qwen2-VL exhibits the opposite behavior with reduced sensitivity, possibly due to its vision-centered architecture enhancing robustness to disturbances in image sequences. In the mixed-modality scenario, the sensitivity of most models is further amplified. This result indicates that the cross-modal attention mechanism in the multimodal interaction process will further magnify the existing positional bias, resulting in the generation performance being more sensitive to the order of evidence arrangement. In terms of the performance differences among models, LLAMa-3.2 shows a relatively higher sensitivity ($PSI\approx 0.045$) under the mixed-modality conditions, suggesting that its cross-modal fusion strategy may be more dependent on the sequence order; The sensitivity of GPT-4o is relatively balanced, indicating that it has a stronger adaptability to positional disturbances. Based on the above findings, in next section, we will further explore the influence mechanism of the retrieval scale on the position bias amplitude through the increment experiment of the retrieval quantity analysis.

\subsection{Bias under Varying Retrieval Sizes}

\begin{figure}[t!]
\begin{center}
\includegraphics[width=1\linewidth]{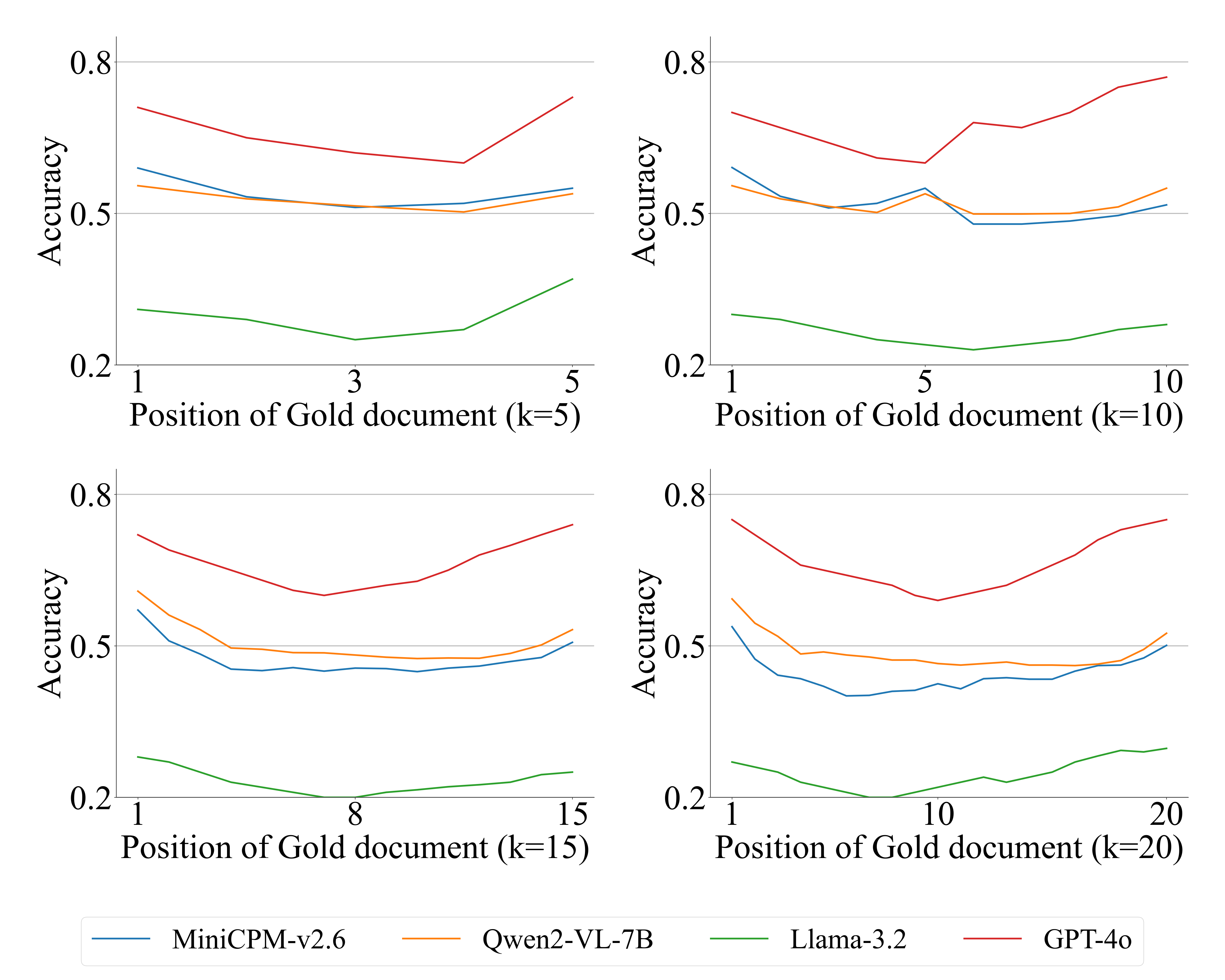} 
\end{center}
\caption{Accuracy curves of the above four models on the ChartQA dataset under varying retrieval sizes $k \in \{5,10,15,20\}$.}
\label{fig:scaling_all} 
\end{figure}

To further explore the robustness and causes of the positional bias phenomenon, in this section, multiple random experiments were conducted on the above four models on the ChartQA dataset. The four subgraphs in Figure \ref{fig:scaling_all} respectively show the corresponding expanded U-shaped curves at \(k\in\{5,10,15,20\}\). For each \(k\), we successively place the correct evidence at position $i$ ($i\in k$), record and plot the average accuracy curve of each model. As the number of retrieval results increases, the accuracy rate at the middle position continuously decreases, while the changes at the top and end positions are relatively small, thereby further magnification the position bias.

Subsequently, we calculated $PSI_p$ based on the average accuracy rate of each position and the sample variance. The results are shown in the Table \ref{tab:kExp}. As shown in the table, the \(PSI_p\) of all models shows a continuous growth trend with the increase of the retrieval quantity \(k\) : When \(k\) increased from 3 to 20, the sensitivity of MINICPM-V2.6 rose from approximately 0.018 to 0.060, and Qwen2-VL increased from 0.020 to 0.080. Llama-3.2 and GPT-4o also increased from 0.030 and 0.028 to 0.085 and 0.082 respectively. The slight increase in the variance term indicates that the results of multiple random experiments are stable and reliable. This phenomenon intuitively reflects that as the number of retrieval results increases, the model neglects the evidence in the middle more seriously. However, the top and end position can still maintain a relatively high level of attention, resulting in a further widening of the performance gap between the different positions, and the model's bias from the sequence order becomes more severe.

\begin{table}[htbp]
\centering
\small
\renewcommand{\arraystretch}{1.25}
\adjustbox{max width=\linewidth}{
\begin{tabular}{c|cccc}
\toprule
\midrule
\(k\) & \textbf{MiniCPM-v2.6} & \textbf{Qwen2-VL-7B} & \textbf{Llama-3.2-11B} & \textbf{GPT-4o} \\
\midrule
 3  & $\mathbf{0.040}{\scriptstyle( \pm 0.012 )}$ & $\mathbf{0.032}{\scriptstyle( \pm 0.010 )}$ & $\mathbf{0.024}{\scriptstyle( \pm 0.011 )}$ & $\mathbf{0.107}{\scriptstyle( \pm 0.009 )}$ \\
 5  & $\mathbf{0.078}{\scriptstyle( \pm 0.009 )}$ & $\mathbf{0.052}{\scriptstyle( \pm 0.013 )}$ & $\mathbf{0.065}{\scriptstyle( \pm 0.010 )}$ & $\mathbf{0.113}{\scriptstyle( \pm 0.010 )}$ \\
10  & $\mathbf{0.094}{\scriptstyle( \pm 0.010 )}$ & $\mathbf{0.114}{\scriptstyle( \pm 0.009 )}$ & $\mathbf{0.097}{\scriptstyle( \pm 0.011 )}$ & $\mathbf{0.119}{\scriptstyle( \pm 0.011 )}$ \\
15  & $\mathbf{0.108}{\scriptstyle( \pm 0.012 )}$ & $\mathbf{0.134}{\scriptstyle( \pm 0.011 )}$ & $\mathbf{0.105}{\scriptstyle( \pm 0.009 )}$ & $\mathbf{0.137}{\scriptstyle( \pm 0.012 )}$ \\
20  & $\mathbf{0.119}{\scriptstyle( \pm 0.013 )}$ & $\mathbf{0.137}{\scriptstyle( \pm 0.012 )}$ & $\mathbf{0.118}{\scriptstyle( \pm 0.014 )}$ & $\mathbf{0.152}{\scriptstyle( \pm 0.013 )}$ \\
\bottomrule
\end{tabular}
}
\caption{Position sensitivity \(PSI_p\) (mean ± variance) of each model under different retrieval quantities \(k\) on ChartQA.}
\label{tab:kExp}
\end{table}

Finally, to characterize the rate at which the degree of bias increases with $k$, we fit the relationship of each model \(PSI_p\) with respect to \(ln(k)\) to a linear model,
\begin{equation}
\begin{aligned}
PSI_p \;=\; \alpha\,\ln(k)\;+\;\beta.
\end{aligned}
\end{equation}

With Llama-3.2, for example, the least squares fitting get \( \ alpha = 0.035 ,  \ beta = 0.010\), goodness-of-fit \(R ^ 2 = 0.986 \). Similarly, the slopes \(\alpha\) of the four models are all between 0.030 and 0.040, verifying the law that the position bias linearly amplifies with the logarithmic growth of the retrieval scale. This result suggests that when designing a multimodal RAG system, the excessive number of retrieval items should be carefully controlled or a position reweighting mechanism should be introduced to alleviate the significant sequence bias caused by the increase of \(k\).

\section{In-depth Exploration of Hidden Bias}
\subsection{Visualization Methodology}

Our study investigates the causes of position bias in multimodal RAG systems via a multi-level visual analysis framework based on the Qwen2-VL-7B and Llama-3.2-11B-Vision-Instruct models. The framework integrates cross-modal attention heatmap visualization and quantitative bias metrics to systematically analyze attention behaviors. The methodology is organized as follows.

Under a fixed retrieval setting $k=3$, we construct controlled sequences by placing the correct evidence at three positions within the retrieved inputs: top (position 1), middle (position 2), and end (position 3). All decoder parameters remain frozen to ensure consistent decoding across configurations. Cross-modal attention weights are extracted from the 14th decoder layer—empirically selected as optimal (justified in the next section)—and rendered as interpretable visualizations.

All input images are preprocessed using bilinear interpolation to a unified resolution of $616\times644$ to ensure spatial consistency. Text tokens are generated from templated prompts and tokenized with the model's native tokenizer. For visual inputs, a Vision Transformer backbone is employed to divide images into spatial patches:
\begin{equation}
T_v = \text{PatchEmbed}(I) \in {R}^{(H/28)\times(W/28)\times d},
\end{equation}
where \(I\) denotes the input image, \(\text{PatchEmbed}\) represents the patch embedding operation, and \(H\), \(W\), \(d\) correspond to the image height, width, and embedding dimension, respectively. The resulting tensor \(T_v\) represents the visual tokens arranged in a 2D spatial grid.

Once the visual and textual tokens are obtained, we extract cross-modal attention maps from key text tokens (e.g., entities or numerals) to the spatial grid of image tokens. To enhance interpretability, the raw attention matrix is normalized within a defined Region of Interest (ROI):
\begin{equation}
A_{\text{norm}}(i, j) = \frac{A(i, j) - \min(A_{\text{ROI}})}{\max(A_{\text{ROI}}) - \min(A_{\text{ROI}})},
\end{equation}
where \(A(i, j)\) is the attention weight at coordinate \((i, j)\), and \(A_{\text{ROI}}\) denotes the subregion associated with projected keyword relevance. The normalized attention maps are then converted into visual overlays. Specifically, we compute an overlay \(\mathcal{O}^{(i)}\) via a visualization mapping function \(\mathcal{M}\) applied to the attention distributions:
\begin{equation}
\mathcal{O}^{(i)} = \mathcal{M}(\alpha_{\text{src}}^{(i)}, \alpha_{\text{tgt}}^{(i)}),
\end{equation}
where \(\mathcal{M}\) maps normalized attention scores from white (low intensity) to red (high intensity). The heatmap is blended with the original image using transparency-based overlaying with a fixed alpha value \((\alpha = 0.5)\):
\begin{equation}
\text{Overlay} = \alpha I + (1 - \alpha)\cdot \text{Heatmap},
\end{equation}
preserving both semantic content and attention saliency.

To support quantitative analysis of positional bias, we further generate “attention difference heatmaps.” These visualizations highlight contrastive patterns in attention allocation. Additionally, a position bias matrix is computed:
\begin{equation}
\Delta A_{\text{pos}} \in {R}^{3\times3},
\end{equation}
along with region-specific attention scores:
\begin{equation}
S_{\text{ROI}} = \frac{1}{N_{\text{patch}}} \sum_{(x, y)\in \text{ROI}} A_{\text{norm}}(x, y),
\end{equation}
and global attention scores:
\begin{equation}
S_{\text{global}} = \frac{1}{HW} \sum_{x = 1}^{H} \sum_{y = 1}^{W} A_{\text{norm}}(x, y),
\end{equation}
to quantify the model's sensitivity to evidence position. Final visualizations are rendered via Matplotlib’s subplot interface for comparative, multi-view presentation.

\subsection{Layer-Specific Sparsity Analysis}
\begin{figure*}[t]
\begin{center}
\includegraphics[width=1\linewidth]{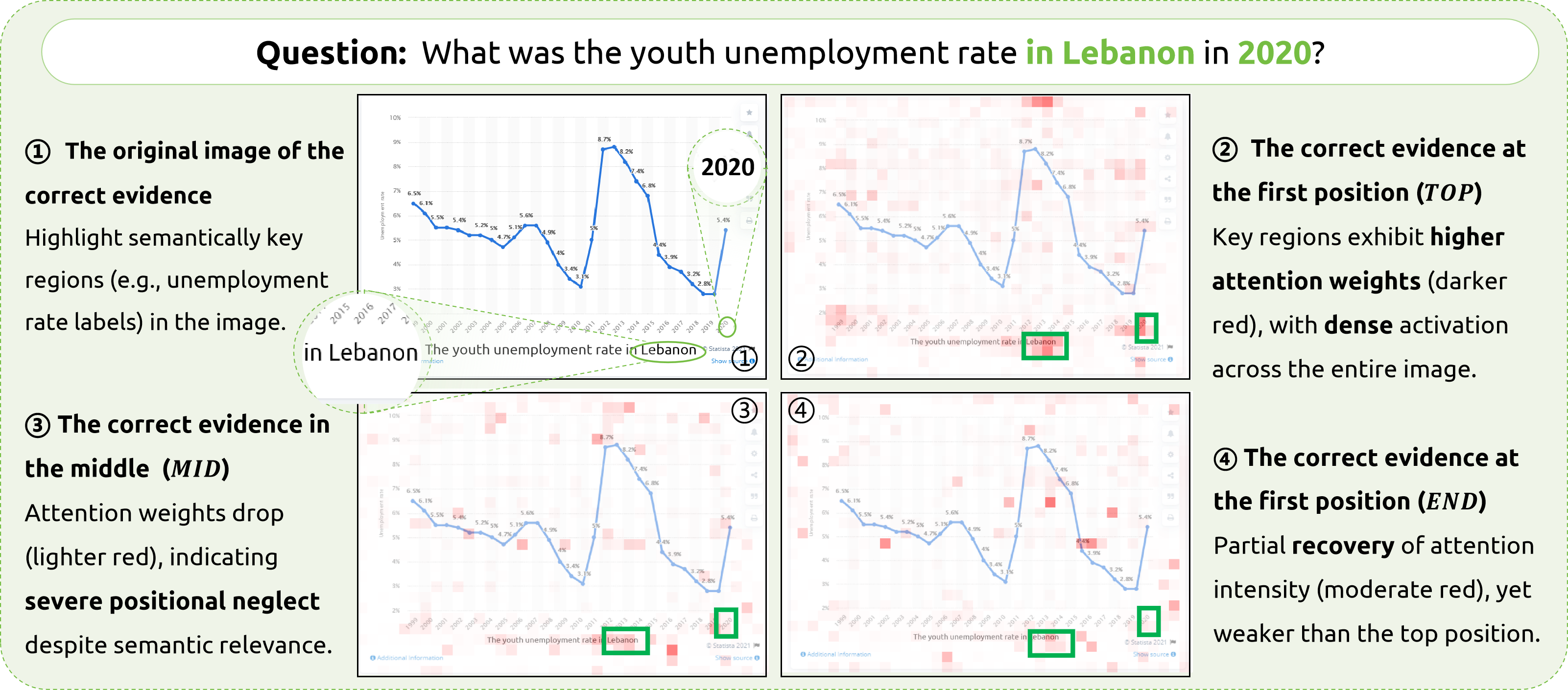} 
\end{center}
\caption{An example of cross-modal attention heatmaps under different evidence positions.}
\label{fig:attn} 
\end{figure*}

To quantify the sparsity degree of cross-modal attention distribution in different decoding layers, we adopt the 'Normalized 2D Entropy' (N2E) index. The indicator definition is as follows: Given a cross-modal attention weight matrix of a certain layer \(\displaystyle A\in R^{H\times W}\), first normalize it:
\begin{equation}
\begin{aligned}
p_{ij} \;=\;\frac{A_{ij}}{\sum_{m=1}^{H}\sum_{n=1}^{W}A_{mn}}.
\end{aligned}
\end{equation}
Next, we compute the two-dimensional Shannon entropy:
\begin{equation}
\begin{aligned}
H_{\!2D}(A)\;=\;-\,\sum_{i=1}^{H}\sum_{j=1}^{W}p_{ij}\,\ln p_{ij}.
\end{aligned}
\end{equation}
This entropy is then normalized to the range \([0,1]\) :
\begin{equation}
\begin{aligned}
\mathrm{N2E}(A)
\;=\;
\frac{H_{\!2D}(A)}{\ln(H\,W)}\,.
\end{aligned}
\end{equation}
Finally, the two-dimensional sparsity index is derived as:
\begin{equation}
\begin{aligned}
S_{\!2D}(A)\;=\;1-\mathrm{N2E}(A).
\end{aligned}
\end{equation}
Among them, \(S_{\!) 2D}=1\) indicates extreme sparsity (attention is highly concentrated on certain patches), \(S_{\! 2D}=0\) indicates a completely uniform distribution.
For Qwen2-VL-7B-instruct and LLama-3.2-11b-vision-instruct, Respectively to extract cross modal cross attention (text and visual) matrix in each layer  \(\ell=0,1,\dots,27\) , and the calculation results of three typical layers (shallow layer \(\ell=3\), middle layer \(\ell=14\), and deep layer \(\ell=24\)) are taken as shown in the Table \ref{tab:layer}.

\begin{table}[h]
  \centering
  \renewcommand{\arraystretch}{1.10}
  \adjustbox{max width=\columnwidth}{
  \begin{tabular}{c|ccc}
    \toprule
    \midrule
     \thead{\textbf{Model / Hierarchy}} & \thead{$\ell=3$\\ (shallow layer)} & \thead{$\ell=14$\\ (middle layer)} & \thead{$\ell=24$\\ (deep layer)} \\
    \midrule
    \textit{Qwen2‑VL‑7B} & \(0.42\) & \(\mathbf{0.72}\) & \(0.68\) \\
    \textit{Llama‑3.2-11B}    & \(0.45\) & \(\mathbf{0.75}\) & \(0.70\) \\
    \bottomrule
  \end{tabular}
  }
  \caption{ \(S_{\!2D}\) at different levels.}
  \label{tab:layer}
  \vspace{-2mm}
\end{table}

From the table, we observe that both models exhibit lowest sparsity in shallow layers, suggesting more uniformly distributed attention. In contrast, middle layers exhibit the highest sparsity, indicating that attention becomes highly concentrated on semantically critical regions, which is a hallmark of effective cross-modal feature alignment. The sparsity remains relatively high in deeper layers, likely reflecting the model’s focus on reasoning-critical visual regions in support of text generation. Based on this quantitative trend, we hypothesize that in the 28-layer decoder of Qwen2-VL, the shallow layers ($\ell=0\sim11$) mainly focus on unimodal processing (e.g., text self-attention or local visual features), the middle layers ($\ell=12\sim19$) progressively enhance cross-modal attention and are critical for cross-modal integration, and the deep layers ($\ell=20\sim27$) primarily engage in text-centric reasoning. Therefore, in the following sections, we select the 14th layer for attention heatmap visualization and position bias analysis, as it represents the point of strongest cross-modal interaction.

\subsection{Attention Discrepancy and Visualization}
\begin{figure*}[t]
\begin{center}
\includegraphics[width=1\linewidth]{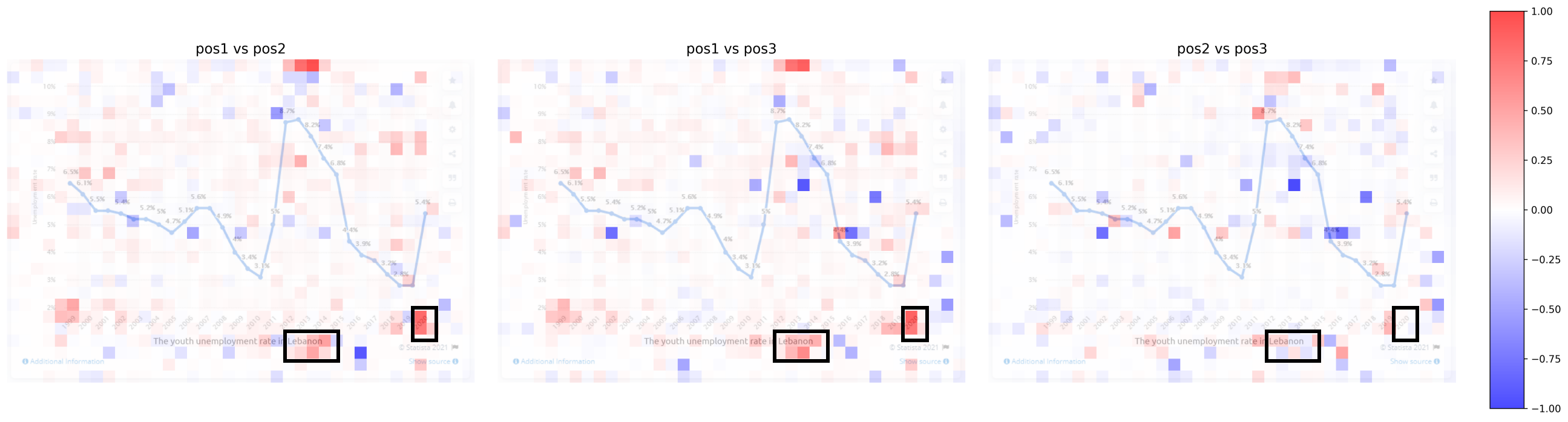} 
\end{center}
\caption{Visualization of attention differences at layer 14 under different gold document position configurations. Each subfigure shows patch-wise attention differences between two settings.  The red-blue colormap indicates relative attention (red = higher in the first configuration; blue = higher in the second). Black boxes highlight semantically relevant regions with notable shifts.}
\label{fig:attn_diff} 
\end{figure*}
\vspace{-1em}

To further investigate the position bias in cross-modal attention, we visualize the attention maps at the 14th decoder layer under three document position configurations: \textit{Top}, \textit{Mid}, and \textit{End}. Figure \ref{fig:attn} shows the semi-transparent overlays of the model’s attention heatmaps overlaid on the corresponding input images, where only one image contains the correct evidence, and the other two serve as distractors. A concrete example is illustrated in Figure \ref{fig:attn}: When analyzing the youth unemployment rate in Lebanon (2020), the model focuses on semantically critical regions such as the numeric labels and axis titles in the bar chart (highlighted by black boxes). Across all cases, we observe that the model consistently attends to semantically relevant regions in the correct image, confirming that the 14th layer captures meaningful cross-modal interactions. However, the magnitude and focus of attention vary noticeably depending on the position of the correct evidence in the retrieval sequence.

To quantitatively examine how attention varies with position, we compute patch-wise differences between attention maps across configurations and visualize them in Figure \ref{fig:attn_diff}. Each heatmap illustrates the difference between two positional settings (Top-Mid, Top-End, and Mid-End), where red indicates increased attention and blue indicates reduced attention for the first position relative to the second.

It can be observed from the difference heatmaps that when the correct document is located at the beginning position, the overall attention received is higher than when it is located in the middle and the end, and the attention received by the local key areas is significantly higher than that at the other two positions. This emphasizes the strong bias of the model towards the main evidence. Moreover, when the correct evidence is located at the end, the overall attention score is also higher than when it is located in the middle position, which confirms that the model tends to underestimate the middle evidence in its cross-modal reasoning.

In summary, the attention heatmaps and their pairwise differences jointly confirm a position-sensitive bias in cross-modal attention allocation. The \textit{Top} and \textit{End} positions induce stronger focus on relevant evidence, while the \textit{Mid} position leads to diluted attention. These findings provide both visual and quantitative support for incorporating position-aware strategies such as evidence reweighting or positional prompts to mitigate attention imbalance.

\section{Related Work}
\paragraph{Single-mode RAG position bias}
Position bias is a known challenge in single-mode Retrieval-Augmented Generation (RAG)\cite{wang2024primacyeffectchatgpt}.
LLMs often exhibit sensitivity to information position in long contexts, with higher attention at sequence ends and neglected middles, a phenomenon linked to pre-training preferences in Transformer models \cite{coelho-etal-2024-dwell}.
Similarly, graph data serialization can alter LLM perception of topological structures \cite{ge2024graphdescriptiveorderaffect}.
Prompt-based attention direction has been proposed to mitigate this bias \cite{zhang-etal-2024-instruct}.

\paragraph{Multimodal RAG systems}
Advancements in multimodal RAG include Google's ColPali framework for end-to-end document understanding using delayed interaction encoders \cite{faysse2024colpaliefficientdocumentretrieval}.
Other approaches involve multi-agent cooperative retrieval planning for enhanced reasoning robustness and cross-modal allocation-causal reasoning architectures targeting semantic gaps with joint embeddings and hierarchical retrieval, notably in medical image QA.

\paragraph{Position Bias in Multimodal RAG} 
While efforts to quantify position bias include metrics like ListT5 for text retrieval \cite{yoon-etal-2024-listt5} and Landmark Embedding with location-aware objectives \cite{luo-etal-2024-landmark}, current quantitative methods are often limited to text-only tasks or do not readily extend to multimodal scenarios.
In such multimodal contexts, position bias also significantly affects Multimodal LLMs (MLLMs).
Work by \cite{tan2024ordermattersexploringorder} demonstrated that MLLMs exhibit this bias, favoring visual features when placed at sequence beginnings or ends.
Although this study confirmed position bias in RAG settings and suggested prompting for mitigation, its experimental design involved a limited number of candidates (e.g., four items).
This setup may not capture potential attention dilution effects present in real-world scenarios involving larger-scale retrieval results (e.g., more than five candidates), a limitation our work addresses.

\section{Conclusion}
Our study establishes foundational insights into position bias in multimodal retrieval-augmented generation (RAG) systems, uncovering systematic performance instability rooted in the ordering of retrieved evidence. Our experiments reveal that multimodal reasoning exhibits heightened sensitivity to positional arrangements compared to unimodal settings, with accuracy following a distinct U-shaped trajectory as retrieval scope expands. To quantify the bias, we introduce a position sensitivity metric $PSI_p$ and an interpretable diagnostic framework, demonstrating that cross-modal attention mechanisms disproportionately prioritize sequence extremities while neglecting middle content. Our work provides a fairer evaluation framework and theoretical support for the design of multimodal RAG systems.

\section*{Limitations}
While our study establishes foundational insights into position bias in multimodal RAG systems, several directions warrant further exploration. First, the experimental scope is constrained to retrieval scales $k\leq20$ and models under 11B parameters due to computational resource limitations(requiring 16+ GPUs), leaving larger-scale scenarios and frontier architectures (e.g. Claude 3) for future investigation. Second, while we empirically observe model-specific sensitivity patterns (e.g. Llama-3.2's higher bias), the analysis does not systematically correlate these differences with architectural designs like attention mechanisms and cross-modal fusion strategies, which could deepen theoretical understanding. Third, although attention visualization reveals critical bias propagation patterns, a unified theoretical framework explaining how positional encoding interacts with multimodal fusion remains to be developed. Fourth, our analysis focuses on static attention distributions, whereas temporal dynamics in multi-step reasoning tasks, where positional effects may evolve, require dedicated study. Finally, operationalizing real-time debiasing mechanisms (e.g. dynamic position reweighting) in practical systems presents an open engineering challenge. These limitations collectively outline promising avenues for advancing both the theory and application of robust multimodal RAG systems.

\section*{Ethical Statement}
This research does not involve human subjects, personally identifiable information, or sensitive data. All experiments are conducted on publicly available benchmark datasets and models (e.g., Qwen2-VL-7B, LLaMA-3.2-11B-Vision-Instruct). We strictly follow the terms of use and licensing agreements of these resources. Our analysis aims to improve the robustness and fairness of multimodal retrieval-augmented generation systems by identifying and mitigating potential positional biases. We believe that our findings can contribute to building more interpretable and responsible AI systems.

\section*{Acknowledgements}
This work is supported in part by the National Key R\&D Program of China under Grant Nos. 2023YFA1011602 and 2023YFC3305303, and the National Natural Science Foundation of China under Grant Nos. 62472408, 62372431, 62441229, and 62377043.

\bibliography{custom}

\appendix

\end{document}